\def\year{2018}\relax
\documentclass[letterpaper]{article} 
\usepackage{aaai18}  
\usepackage{times}  
\usepackage{helvet}  
\usepackage{courier}  
\usepackage{url}  
\usepackage{graphicx}  
\usepackage{amsmath}
\usepackage{amssymb}
\usepackage{algorithm}
\usepackage[noend]{algpseudocode}
\usepackage{array}
\usepackage{csquotes}
\usepackage{subfig}
\begin{document}
%
\title{Q-Learning with Differential Entropy of Q-Tables}
\author{Tung D. Nguyen, Kathryn E. Kasmarik, Hussein A. Abbass\\
The University of New South Wales - Canberra\\
Northcott Drive\\
Canberra BC, ACT 2600\\
}
\maketitle
\begin{abstract}

It is well-known that information loss can occur in the classic and simple Q-learning algorithm. Entropy-based policy search methods were introduced to replace Q-learning and to design algorithms that are more robust against information loss. We conjecture that the reduction in performance during prolonged training sessions of Q-learning is caused by a loss of information, which is non-transparent when only examining the cumulative reward without changing the Q-learning algorithm itself. We introduce Differential Entropy of Q-tables (DE-QT) as an external information loss detector to the Q-learning algorithm. The behaviour of DE-QT over training episodes is analyzed to find an appropriate stopping criterion during training. The results reveal that DE-QT can detect the most appropriate stopping point, where a balance between a high success rate and a high efficiency is met for classic Q-Learning algorithm. 

\end{abstract}

\section{Introduction}

A reinforcement learning (RL) agent searches for the best policy that maps the agent\textquoteright s state space to the action
space while maximizing a total reward function~\cite{sutton1998reinforcement}. The reward function acts as a guiding signal for the agent during learning to filter out less-promising state-action pairs from more promising ones. RL has been used in a wide range of applications including robot skill bootstrapping and human-robot interaction~\cite{kormushev2010robot,mitsunaga2008adapting}. 

The design of an RL agent involves decisions on four components: a suitable representation and definition of the state space, a suitable choice of a set of actions of relevance to the problem at hand, the design of an appropriate reward function to create a gradient for the model to converge to optimal behavior, and a representation of the mapping function that maps states to actions.

The first and fourth components, state-space representation and state-action mapping representation, are the focus of this paper. The tight coupling between these two components has meant that a decision on one influences the other; but more importantly influences the ability of the agent to learn. What sometimes appears to be a simple change to the definition of the state-space can have a dramatic increase in the complexity of the search space. Meanwhile, in addition to the designer\textquoteright s expertise with RL, the agent\textquoteright s performance on a task as measured by accumulated rewards has been the main criterion used to judge the success or otherwise of an RL agent.

Previous studies on Q-learning~\cite{mannor2003cross,araabi2007study,peters2010relative} replaced the algorithm with different forms of entropy-based search algorithms. The analysis implicated that the gradient of the classic Q-learning causes information loss. A different perspective we take in this work is to show that with an appropriate stopping criterion that can detect information loss when it occurs, classic Q-learning, which uses simpler and faster update than more complex RL systems, can still be used for training the agent. 

In this paper, we transform the episodes during training an RL agent to a time series using differential entropy of Q-tables (DE-QT). This time series represents the dynamics of the amount of valuable information that the agent learns with a Q-learning algorithm over time. Based on the behaviour of the time series, we choose the Q-tables at some critical time point in the training session to implement the testing phase. The performance of the RL agent is then analyzed to reveal some interesting patterns. 

The analysis associates the differential entropy of Q-tables with the success of the Q-learning algorithm on the final achievement of a generalized policy towards dynamic environments. The analysis attempts to serve as a criterion for selecting an appropriate stopping point in training session of Q-learning algorithm.

The next section introduces fundamental background information on how the state space has been represented in previous studies and on the conventional Q-Learning algorithm. Next, differential entropy is then introduced as an indicator for detecting the success or failure of the use of different state space representations. The indicator extracted from Q-tables during a training session is then evaluated by the implementation of a dynamic flag collection task before the paper is concluded in the final section.

\section{Background}

Depending on the problem, the state space of the environment can be discrete or continuous, requiring different treatments. In the simple case where both the state and action spaces are discrete, such as in a navigation task for an agent situated in a gridworld, the value function in an RL algorithm can be represented by a look-up table. For high dimensional state spaces or continuous state spaces, it is essential to decide how fine-grained they need to be represented~\cite{kober2013reinforcement} while accounting for \enquote{curse of dimensionality}. 

Currently, in continuous state problems, there are a broad range of techniques used to partition the continuous state space into smaller areas; each can be represented by a group state for subsequent tabular Q-learning. \citeauthor{mao2012q}~\shortcite{mao2012q} proposed a Q-tree algorithm using a decision tree to form a hierarchical state representation containing \enquote{superior states} for different clusters of observations. Methods such as nearest neighbor can also be used as a state space segmentation method \cite{lee2004adaptive}. Another popular example is QLASS introduced by \citeauthor{murao1997q} \shortcite{murao1997q}. Partitioning in QLASS used reinforcement signals to dynamically change the sub-space representing a collection of sensor vectors. 
Although most related literature offer a wide range of techniques to manually or adaptively construct state spaces, the utility of discrete representations, especially in the case of examining the generalization of Q-learning, is only evaluated by the variation of reward or cumulative reward over the training steps. The performance obtained at the end of the training session does not provide a transparent trace of how the Q-values or the belief of the learning agent evolve over time. 

\subsection{Q-Learning}

A classic reinforcement learning problem can be represented by the fundamental concept of Markov Decision Processes (MDPs) \cite{kaelbling1996reinforcement}. An optimal policy $\pi$ is required to be found through a model containing a set \( \mathcal{S} \) of all possible world states, a set \( \mathcal{A} \) of all possible actions $a$, a reward function \( \mathcal{R}(s,a) \), and a transition model $P(s'|s,a)$ representing the probability of transiting to state $s'$ after taking action $a$ from state $s$. 

Commonly, the objective of a learning agent is to achieve the highest cumulative discounted reward over time $E(\sum_{i=1} \gamma^{i}r_i)$, where $r_i$ is the instant reward received at the $i^{th}$ step, and $\gamma$ is the discount factor whose value is in the range of $[0,1)$. The value of an action $a$ for a state $s$ in this discounted scheme is defined by the value function: 

\begin{equation}
 Q^*(s,a) = R(s,a)+\sum_{s'\in{S}} P(s'|s,a))\gamma V^*(s')
\end{equation}

Q-learning algorithm is a model-free method that attempts to learn Q-values without knowing the transition model $P(s'|s,a)$. The algorithm starts with some initial assumptions on the Q-values for all state-action pairs which are then gradually updated through interactions with the environment before finally converging to $Q^*(s,a)$ \cite{Watkins1992}. 

In reinforcement learning, an action selection method is an important factor as one has to consider the balance between exploration and exploitation. The selection method used in this paper is Boltzmann distribution using softmax function to compute the probabilities of the actions in relation with their Q-values given a state:
\begin{equation}
 P(a)=\frac{e^{Q(s,a)/T}}{\sum_{b\in{\mathcal{A}}} e^{Q(s,b)/T}}
\end{equation} 
where, $T$ is the temperature affecting the randomness of the actions. The higher the value of $T$, the more exploration behaviour the agent performs. The temperature is reduced gradually during experiments to facilitate more greedy actions.

\section{Differential Entropy as an Information Loss Detector}

Differential entropy is used to estimate the amount of information produced by a continuous probability distribution. Given a set $X$ with $M$ observations $X=\{x^{(1)},x^{(2)},...,x^{(M)}\}$, the differential entropy of $X$ with a probability distribution $f(x)$, $x\in\chi$ is then defined as:

\begin{equation}
 \varepsilon{(X)} = -\int_\chi f(x)\log{f(x)}dx
\end{equation}

Due to the absence of an explicit mathematical expression of the real density function in most cases, it is tractable to compute the differential entropy through the use of different density estimators. In this paper, we consider the probability distribution estimated by the most widely used density estimator, the histogram. Denote the histogram estimator as $\hat{f}(x)$, the differential entropy is equivalent to the discrete entropy approximated on the histogram of $n$ bins:

\begin{equation}
 \varepsilon{(X)} = -\sum_{i=1}^{n} \hat{f} (x^{(i)})\log{\frac{\hat{f} (x^{(i)})}{w(x^{(i)})}}
\end{equation}

\subsection{Differential Entropy of Q-Tables (DE-QT)}

This section aims at introducing the differential entropy to the Q-learning context. We relates the DE-QT to information loss over the training session.

Given an M-state problem, the state-action pairs\textquoteright Q-values in an episode $t$ is included in a Q-table expressed as $Q(X_t)=[q(x_t^{(1)}),q(x_t^{(2)}),...,q(x_t^{(M)})]$, where $x_t^{i}$ denotes the state $i$ in the state space. We can define a function $f:\varepsilon{(Q(X_t))} \mapsto J(X_t)$ which maps the value of the differential entropy of a sequence to $J(X_t)$ representing the information value of the Q-table comprehended by the learning agent. Suppose we have a transition vector comprising all state transitions in the episode $t$, $v_t = <\phi_t^{1},\phi_t^{2},...,\phi_t^{K}>$, where $\phi_t^{k}$ is the $k^{th}$ state transition in the episode $t$. There is a proportional relationship between the information value $J(X_t)$ of the Q-table and the performance function that one agent obtains at the end of the training episode:

\begin{equation}
 \mathcal{S}(v_t) = \sum_{k=1}^{K}r_{t}(\phi_t^{i})
\end{equation}
where $r_t$ is the reward obtained at time $t$. Based on the value of the differential entropy, we can find the relevance between the information loss and the performance of Q-table after each training episode. Let us denote the maximum information that one can obtain by $J^*$, then:

\begin{equation}
 J^* = max_{\varepsilon\in\mathcal{E}} J(\varepsilon)
\end{equation}

Thus, the information loss compared to the maximum information is $\delta J_t = J^* - J(\varepsilon_t)$. To avoid information loss due to under-training and over-training problems, the stopping condition for the training session has to minimize the $\delta J_t$ while maintaining a sufficient value of performance $\mathcal{S}(v_t)$. 

\section{Experimental Description}

In the experiments, we use a modified version of the Maze problem, whose original version was introduced by \citeauthor{dearden1998bayesian} \shortcite{dearden1998bayesian}. The experiments aim at investigating the influence of state space representations on the robustness of the Q-learning algorithm in a bounded-dynamic environment.

\subsection{Flag Collection Task}
The idea of Maze domain can be used to analyze numerous RL problems due to the existence of multiple local optima that challenges the exploration-exploitation balance of a learning strategy. However, the original setup for Maze problem was a static scenario:
\begin{itemize}
\item The locations of objects in the environment are fixed.
\item All information included in the state space are available to the agent. The flags are identifiable, i.e the positions and their availability at those positions are transparent to the agent. In other words, the task can be solely defined by a basic Markov Decision Process. 
\end{itemize}

To make the problem more interesting, we include one more level of bounded dynamics allowing the position and the number of flags to vary within a fixed region. Besides, the shortcoming in fully defined state space is that (1) it is unable to generalize in real-world processes, and (2) it can grow exponentially in size. In one representation, the dimensions of the state space is allowed to exponentially grow with an increasing number of flags and the number of available positions for flags. In this paper, we facilitate the generalization of the algorithm by reducing the information encoded in a state space so that the environment is partially observable.

\begin{figure}
  \subfloat[][Flag collection task\label{subfig-1:environment}]{%
       \includegraphics[width=0.2\textwidth]{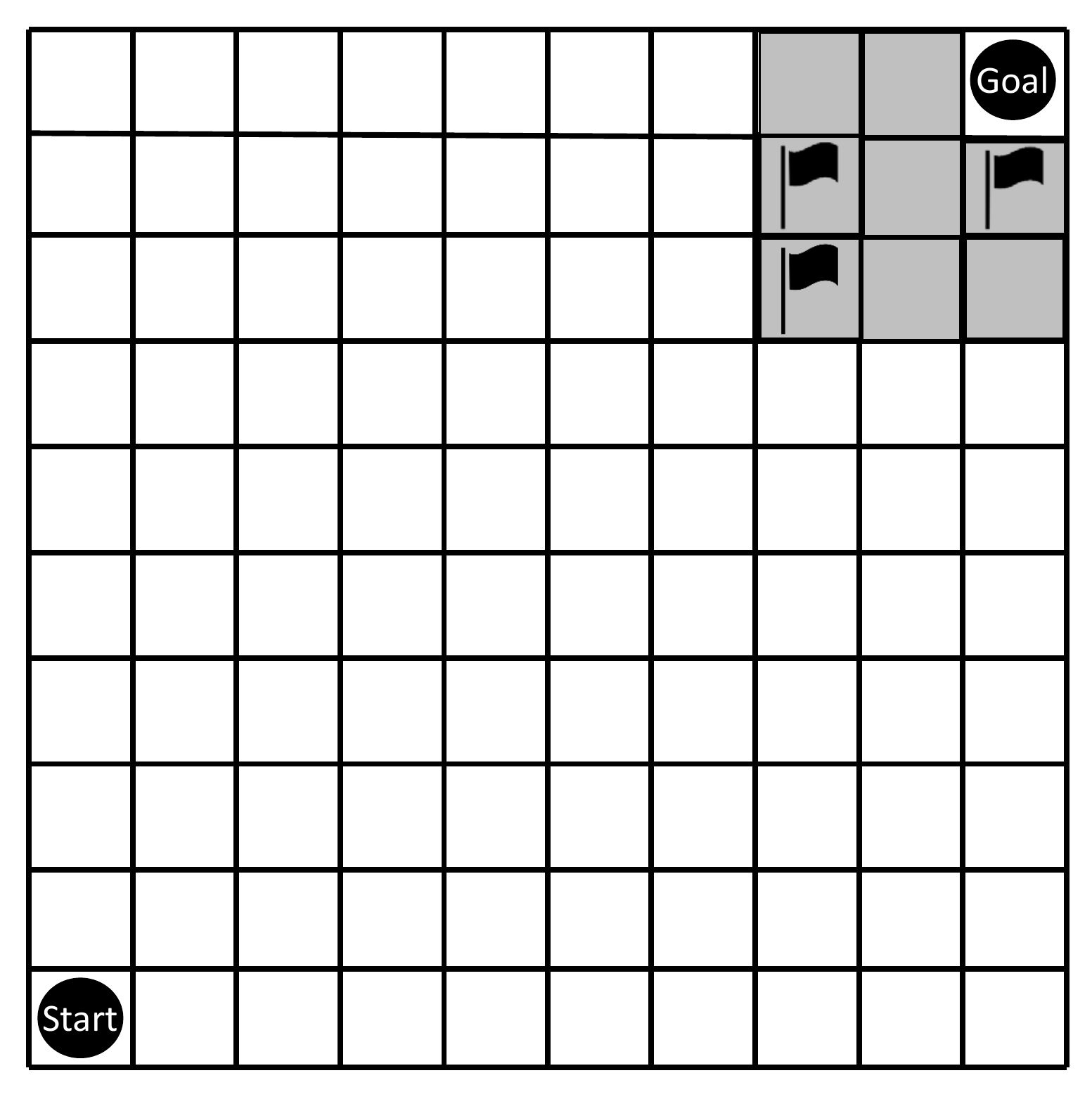}
     }     
     \subfloat[][State space\label{subfig-2:state-space-representation}]{%
       \includegraphics[width=0.25\textwidth]{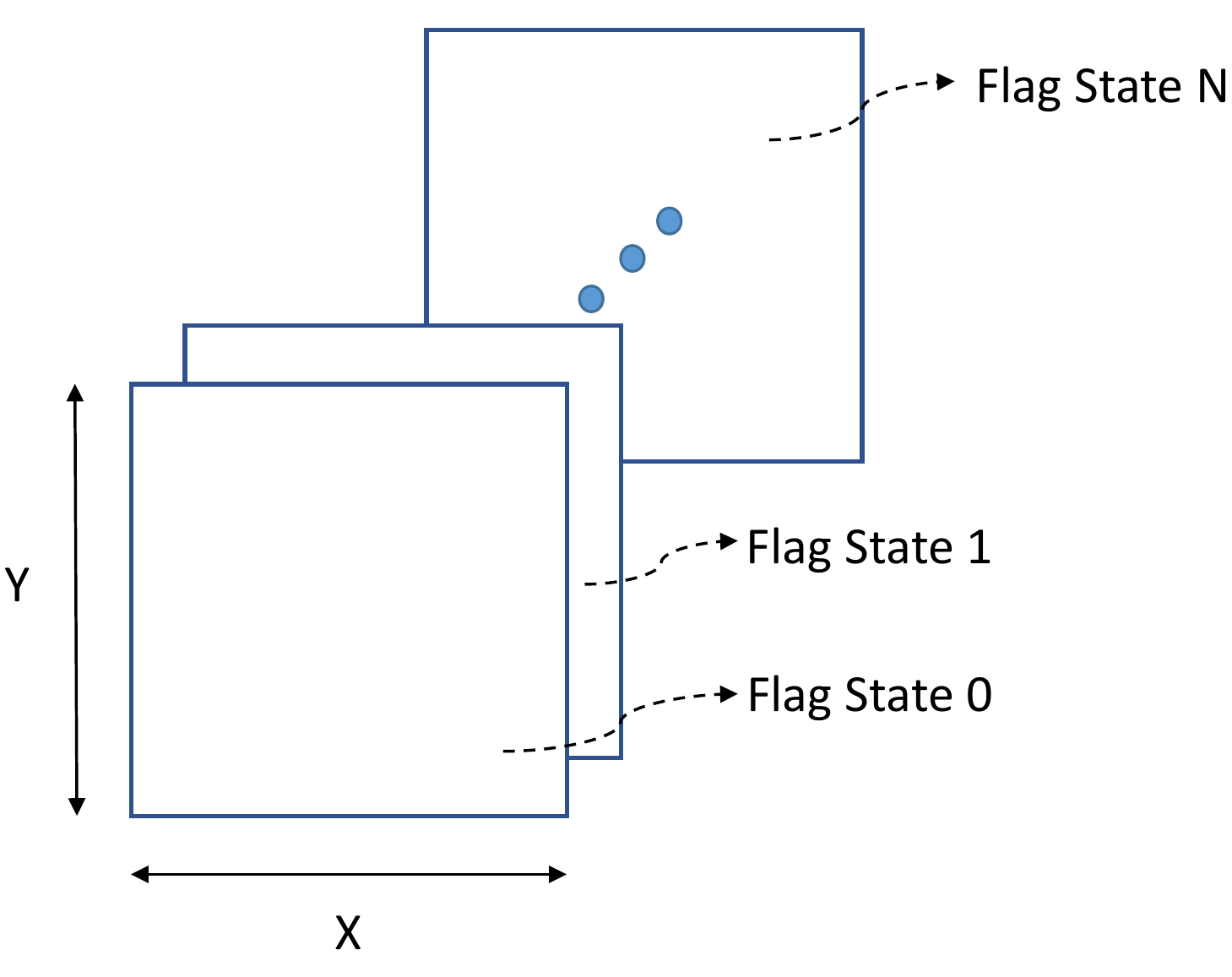}
     }
     \caption{Visualization of environment and state space representation.}
     \label{fig:task}
\end{figure}

Figure~\ref{subfig-1:environment} shows the 10$\times$10 grid world environment where the agent starts at one corner, tries to collect all flags on the way to the opposite corner. The scenario is mimicking a UAV travelling from a base, undertaking a surveillance mission then landing at the other end behind the surveillance area. The shaded area is the area corresponding to the radius of 2 from the goal where flags might be located. The agent only receives an amount of rewards equals to the number of flags collected when it reaches the goal. One episode ends if the agent reaches the goal or it takes 1000 actions (up, down, left, or right). The agent stays at the same position if it attempts to move out-of-bound.

\subsection{State Space Representations}

Figure \ref{subfig-2:state-space-representation} shows the spatial structure of the state space. The state space is defined by three variables:
\begin{itemize}
\item Two \emph{positional variables}: The location of the agent in the world defined by the $X$ and $Y$ variables. The size of the positional states alone is $10\times 10 = 100$ states.
\item \emph{Flag state}: The status of flags to be collected. In the experiments below, different representations of this flag state are used. 
\end{itemize}

The utilization of different state space representations helps provide different insights on how effective the entropy information is in indicating the success of a training session considering a variety of available information.

\subsection{Experiments}
In the experiments described below, the area where the flags are located has a radius of two steps from the goal. This results in 8 possible locations (excluding the goal position) for flags. The number of flags used in training the agent is varied from 1 to 8 flags with different flag state representations. Table~\ref{table:parameters} shows the fixed parameters for the Q-learning algorithm. 

The Q-table of state-action pairs Q values is uniformly initialized to a low value of 0.1. Boltzmann probability distribution is used as the action selection method. Each experimental setup is evaluated by averaging the discounted reward, the number of flags collected (even if the goal is not reached) and the rate of success (collecting enough flags and reaching the goal) over 30 runs.

\begin{table}
\centering
    \begin{tabular}{ l | l}
    \textbf{Terms} & \textbf{Value}\\
    \hline
    Learning rate ($\alpha$) & 0.1 \\ 
    Discount factor ($\gamma$) & 0.999 \\
    Initial temperature ($T_0$) & 1000 \\
    Temperature decay term ($\delta T$) & 0.99 \\
    Temperature updating iteration & 1000 \\
    Minimum temperature & 0.1 \\
    Maximum number of episodes & 10000 \\
    \end{tabular}
    \caption{Learning parameters.}
    \label{table:parameters}
\end{table}

In the experiments below, different flag state representations (the third dimension of the state space) are used. The success of the learning algorithm in this paper is defined differently than it is defined in classic Q-learning, where it is only measured by the cumulative reward. The generalization of the policy is used as a measure of success with which the agent has to learn a policy to  perform the task well in a dynamic environment. We introduce two stages in the experiments: training and testing. The state space is always represented a tensor of $H,W,F$, where $H$ is the height of the environment and equals 10 in this paper, $W$ is the width and equals 10 in this paper, and $F$ is the third dimension that we define in three different ways in this paper. We will call this dimension the flag state. The selected state space representations, and the training and testing procedures are explained as follows: 

\subsubsection{Experiment 1 - Global Flag State Representation}

In experiment 1, with $N$ flags, $F$ is $N+1$. We label this experiment as the \enquote{global flag state representation}. This dimension represents the number of flags that need to be collected. For example, if there are 3 flags to be collected, the state space is $10\times10\times4$. The agent starts from the matrix $F=4$ in the tensor. Flags are indistinguishable; thus, when the agent collects any single flag regardless of which flag it collects, it works with the state space of the matrix $F=3$, and so on. When the agent collects all flags, it works with the state space of the matrix $F=0$ to reach the goal. Thus, with $N$ flags in training, the size of the state space is $10 \times 10 \times (N+1)$, where the state $0$ indicates that all flags are collected and the agent has to learn to proceed to the goal as fast as possible.

In this experiment, there are 8 setups used in training sessions where we train the agent on a varying number of flags from 1 to 8, respectively.  After each episode, the location of the flag is randomized within the assigned radius of 2 from the goal. For convenience, we refer to these setups as \emph{Global-1-8} to \emph{Global-8-8}.

\subsubsection{Experiment 2 - Compact Global Flag State Representation}

The context of this experiment is the same as the previous experiment. However, a more compact version (\emph{3-flag states}) of the encoded flag state representation is used with a $10\times10\times3$ tensor regardless of the number of flags. The flag state is categorized into 3 types: more than 1 flags remaining (encoded by 2), only one last flag remaining (encoded by 1), and no flag remaining (encoded by 0). We train the agent on scenarios with 8 flags located within the radius. Then, the testing phase is performed with 1000 runs.

\subsubsection{Experiment 3 - Local Flag State Representation}

In the same context, we define a totally different representation of the flag state where only local information is available to the agent. The agent only knows the state of the flags at its current position. This results in flag state dimension of 2: 0 - no flag at current location; and 1 - there is a flag at the current location. Thus, the dimension of the state space is fixed to $10\times10\times2$ regardless of the number of flags in the problem.

In training, two representative setups are investigated: one flag (\emph{Local-1-8}) and 8 flags (\emph{Local-8-8}) are randomly positioned within a radius of 2 from the goal for every episode. The testing phase is similar to the testing phase in experiment 2 except that the agent only knows local information about flags.

\subsubsection{Testing}

We introduce a testing phase of 1000 tests on the Q-tables learned by the agent in each setup. In each test, one scenario of 8 flags (flags cover all possible locations within a radius of 2 from the goal) is used to test whether the agent learned a \enquote{general} policy to effectively collect all flags. If all possible flag positions within the radius are reached by the agent, the agent is considered to have learnt the policy to collect all flags. The action is performed with a temperature of 0.1 to facilitate the stochastic process. For testing with global flag state representations in experiment 1, the flag state is maintained equal to the number of flags, $N$, from the start of the run until less than $N$ remaining flags is achieved. For example, if during training we have 3 flags, the state space is $10\times10\times3$. During testing, the agent will remain in the matrix $F=3$ until it collects 5 flags before it transitions to $F=2$.

Two types of testing are performed: (1) testing at the end of the training (after 10000 episodes), and (2) early testing at some critical episodes chosen based on DE-QT. 

\begin{figure*}[hbt]

     \subfloat[][Global-1-8 Setup\label{subfig-1:setup-1-8-entropy}]{\includegraphics[width=0.45\textwidth,height=0.25\textheight]{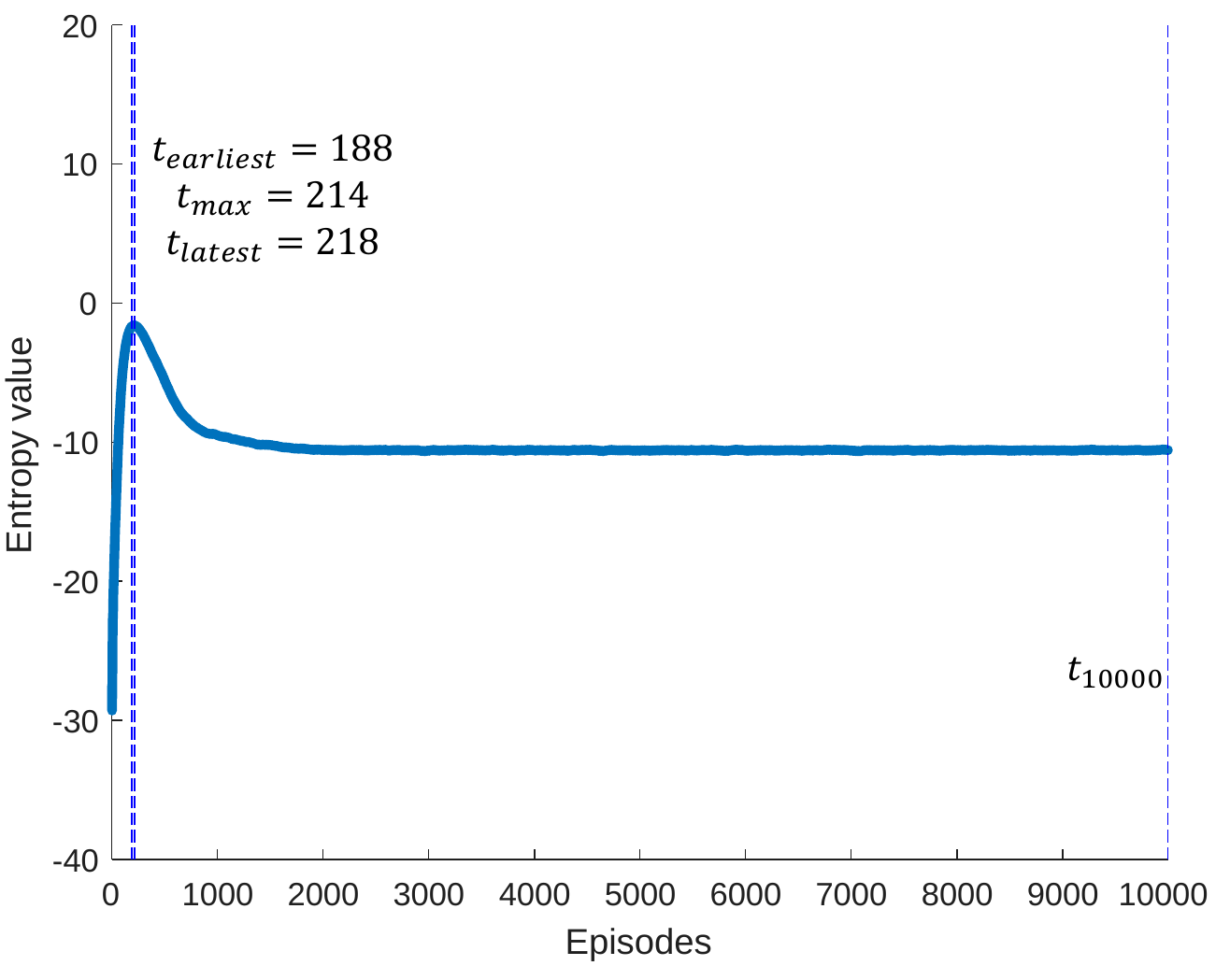}
     }
     \subfloat[][Global-8-8 Setup\label{subfig-1:setup-8-8-entropy}]        {%
       \includegraphics[width=0.45\textwidth,height=0.25\textheight]{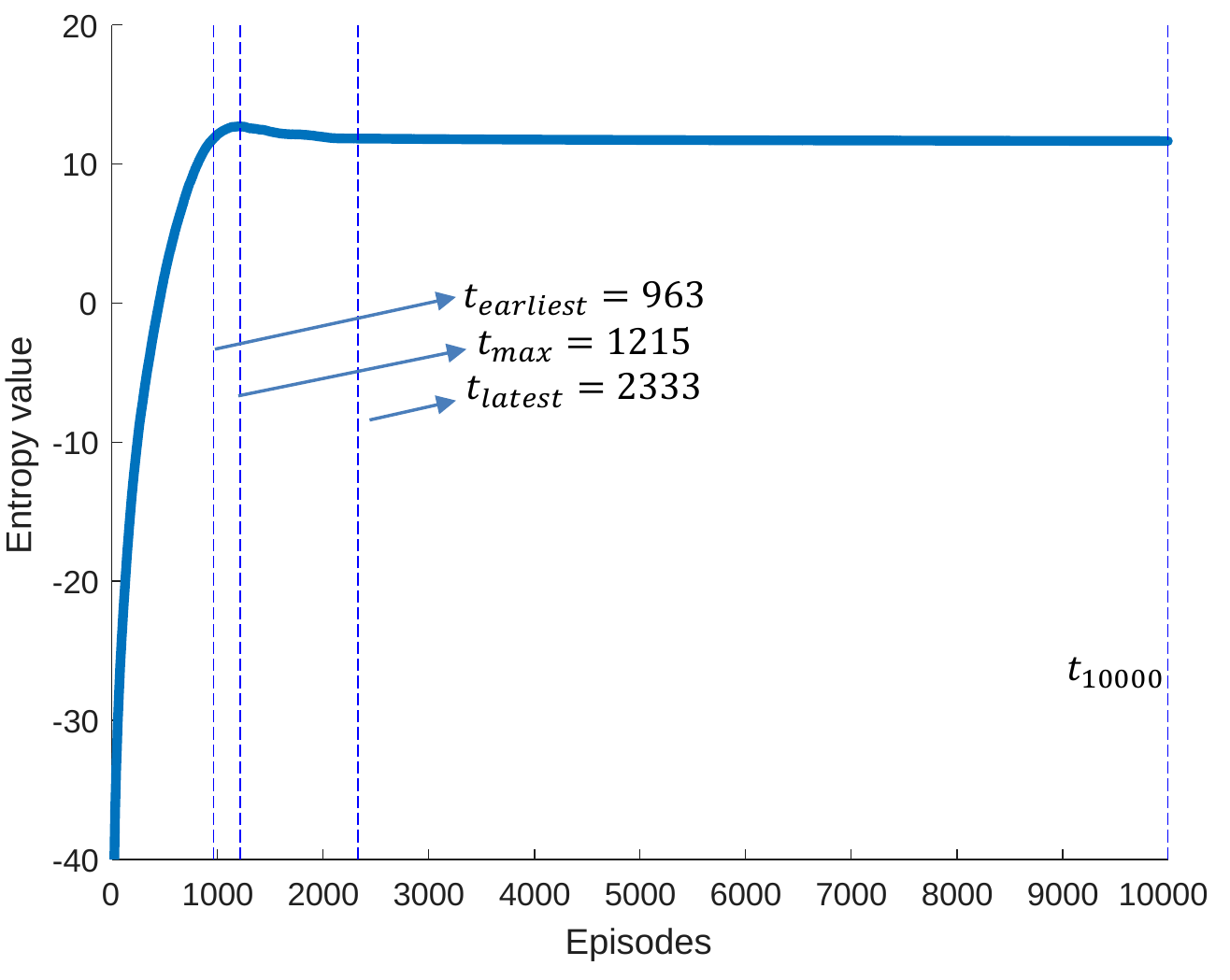}
     }
 
     \subfloat[][Compact State Setup\label{subfig-2:setup-3-state-entropy}]{%
       \includegraphics[width=0.45\textwidth,height=0.25\textheight]{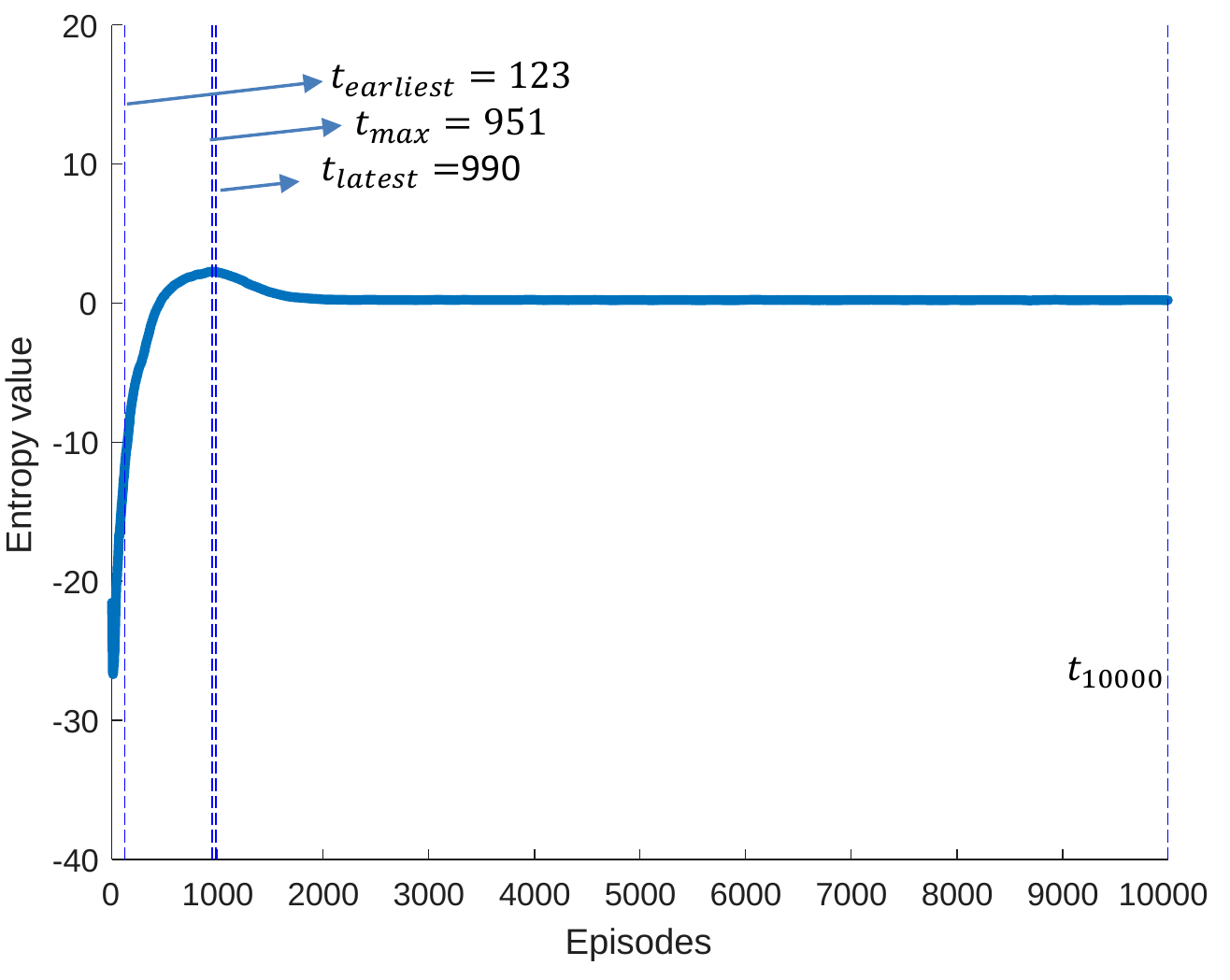}
     }     
     \subfloat[][Local-8-8 Setup\label{subfig-3:setup-localstate-8-8-entropy}]{%
       \includegraphics[width=0.45\textwidth,height=0.25\textheight]{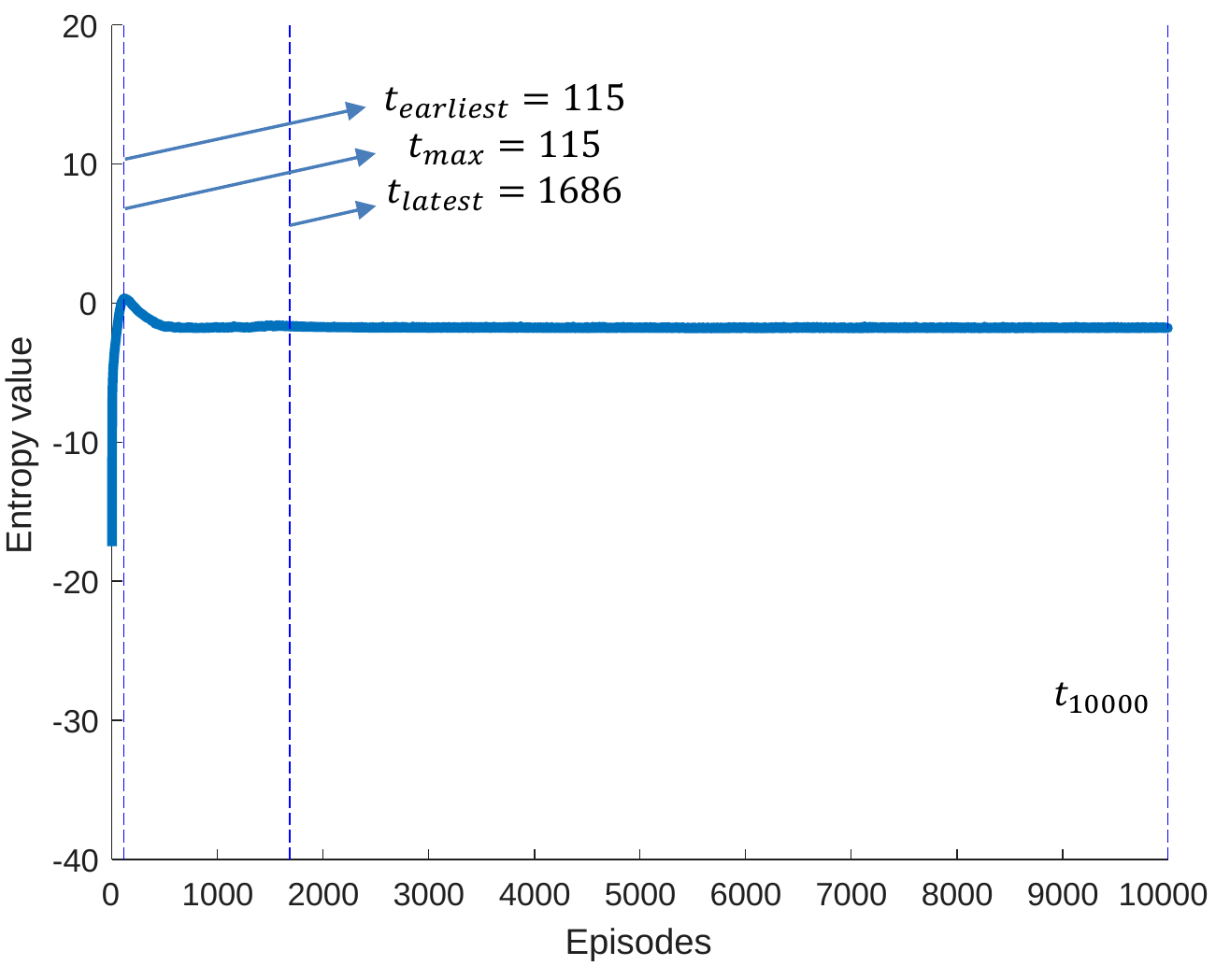}
     }
     \caption{Mean of sum of DE-QT time series of all flag states of four experimental setups over 30 runs.}
     \label{fig:entropy}
\end{figure*}

\begin{figure*}[hbt]

     \subfloat[][Global-1-8 Setup\label{subfig-1:setup-1-8-boxplot}]{\includegraphics[width=0.45\textwidth,height=0.2\textheight]{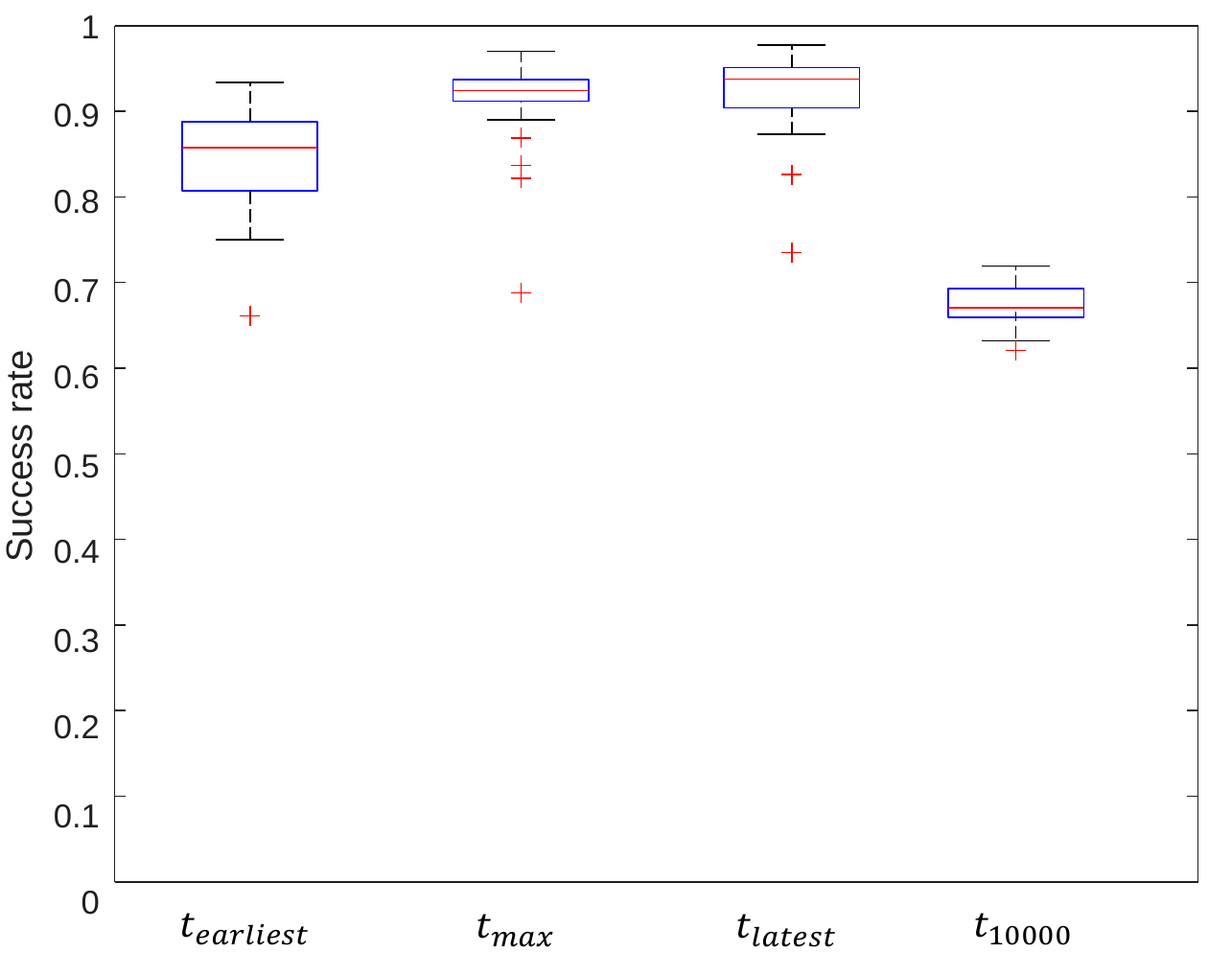}
     }
     \subfloat[][Global-8-8 Setup\label{subfig-1:setup-8-8-boxplot}]        {%
       \includegraphics[width=0.45\textwidth,height=0.2\textheight]{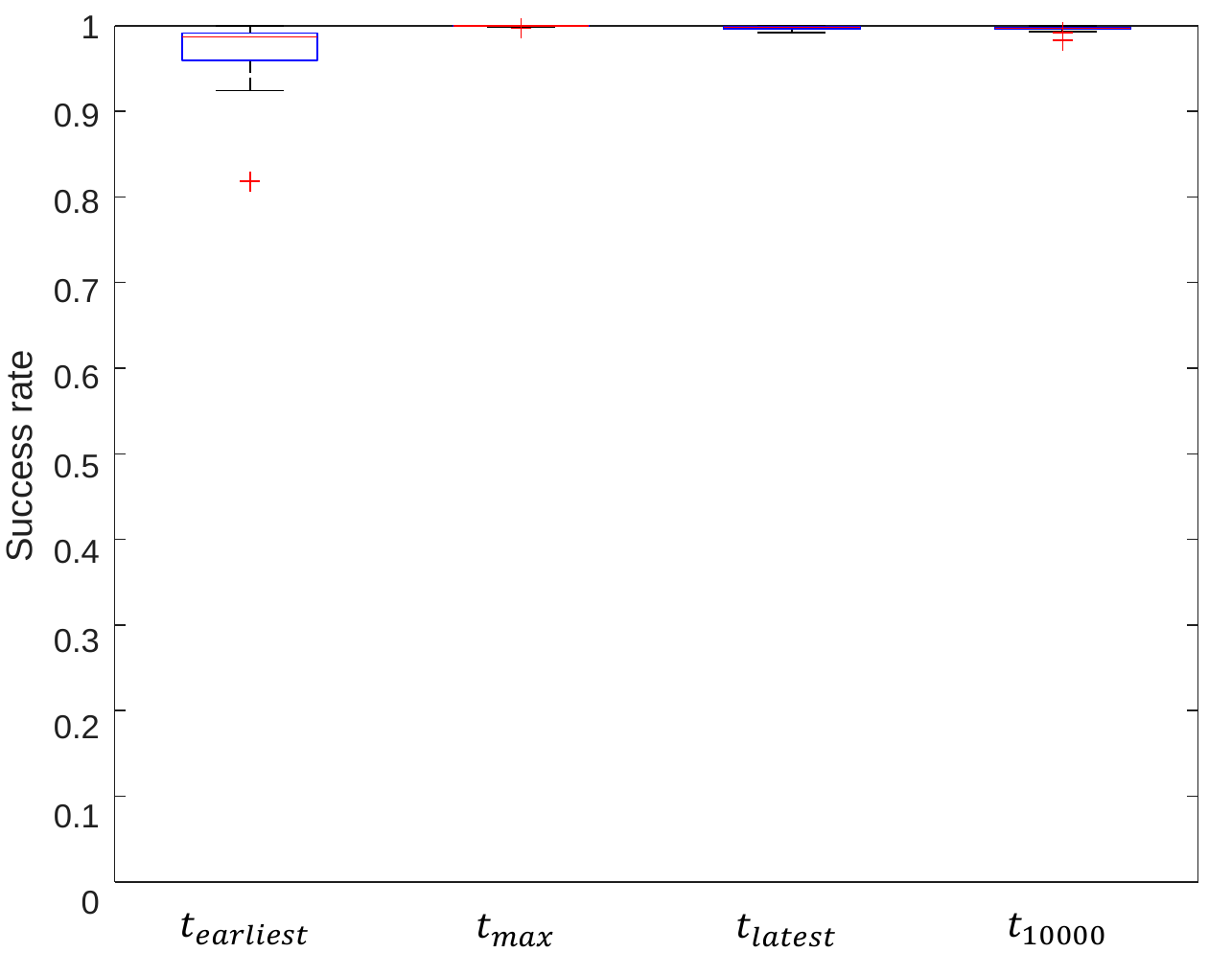}
     }
 
     \subfloat[][Compact State Setup\label{subfig-2:setup-3-state-boxplot}]{%
       \includegraphics[width=0.45\textwidth,height=0.2\textheight]{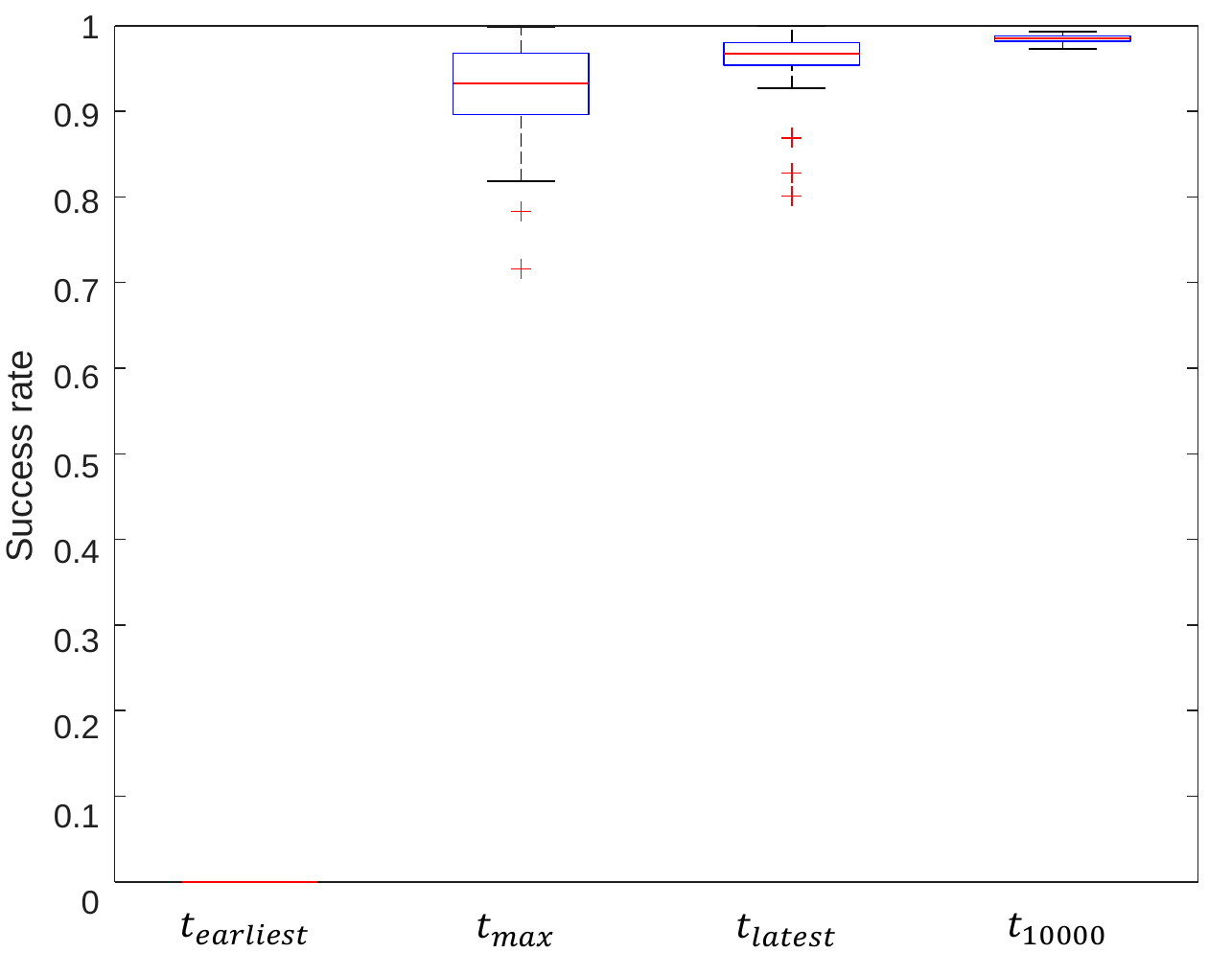}
     }     
     \subfloat[][Local-8-8 Setup\label{subfig-3:setup-localstate-8-8-boxplot}]{%
       \includegraphics[width=0.45\textwidth,height=0.2\textheight]{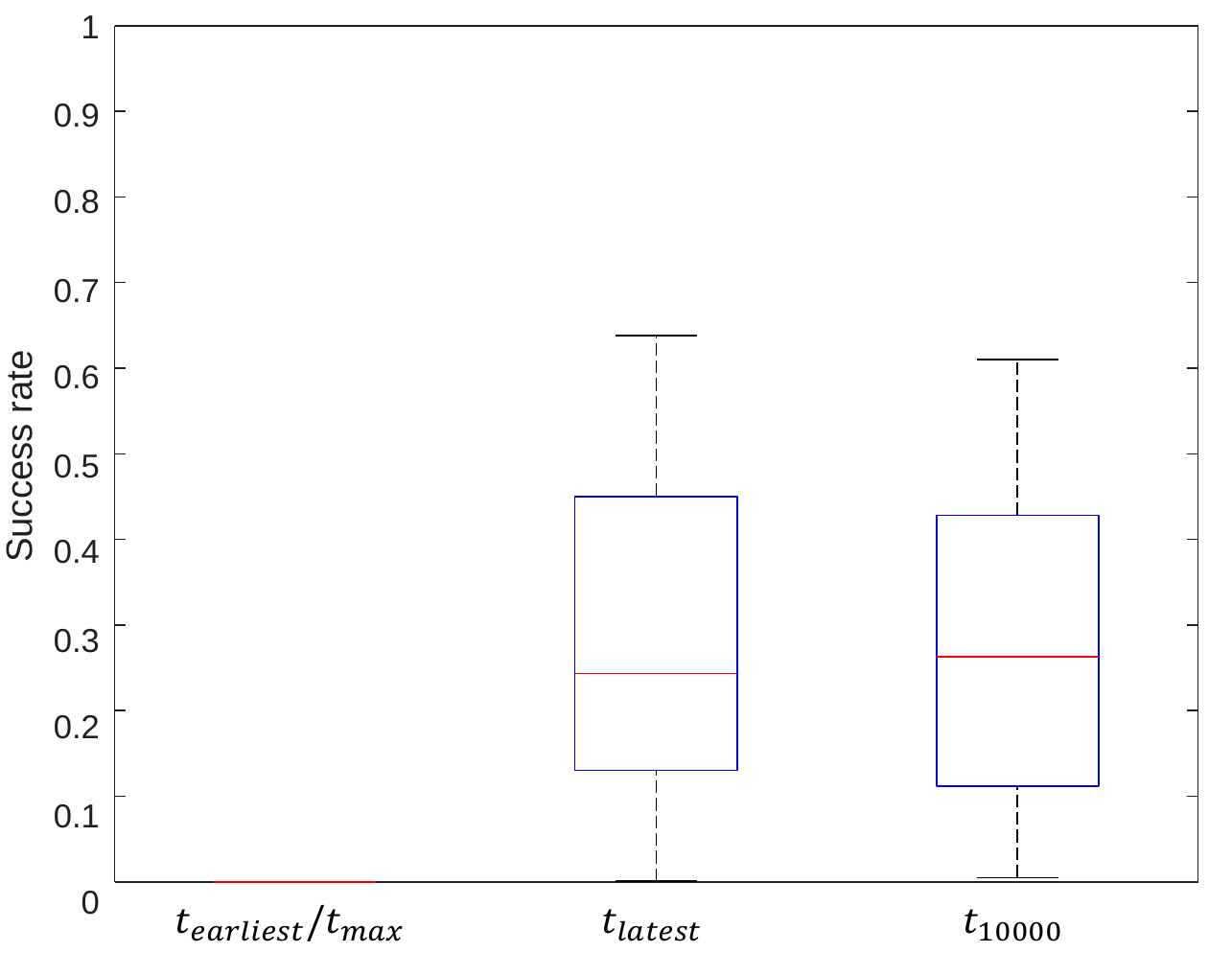}
     }
     \caption{Success rate of four experimental setups at four different testing time over 30 runs.}
     \label{fig:boxplot}
\end{figure*}

\subsection{Workflow}

To assess the performance of Q-tables at different episodes, we introduce a training-testing workflow. 
\begin{itemize}
\item First, the DE-QT estimates the differential entropy values of the Q-tables over episodes. Each flag state corresponds to one channel of the DE-QT time series. We denote the time-series obtained as ${\varepsilon_1(t), \varepsilon_2(t),..., \varepsilon_N(t)}$ for flag state $0$ to flag state $N$.
\item In early testing, among the maximum values of DE-QT series for different flag states, select the Q-tables for testing at the episodes corresponding to the earliest one and the last one:
\begin{equation}
 t_{earliest} = argmin_t\{max(\varepsilon_1(t)),...,max(\varepsilon_N(t))\}
\end{equation}
\begin{equation}
 t_{latest} = argmax_t\{max(\varepsilon_1(t)),...,max(\varepsilon_N(t))\}
\end{equation}
\item Add the series together to create a sum of all differential entropy information on all flag states. Make another test with the Q-table taken at the time of the maximum value of this series ($t_{max}$)
\item Finally, perform post-training test at time $t_{10000}$
\end{itemize} 

\subsection{Generalization evaluation}
The generalization of the training algorithm is tested through the testing phase introduced in the previous section. There are two important factors in generalization of the state space representations: (1) success rate, and (2) efficiency.

One successful test is a test in which the agent collects all 8 flags and proceeds to the goal. The success rate is computed as below:

\begin{equation}
 Success\,Rate = \frac{(Number \, of \, Successful \, Tests)}{1000}
\end{equation}

The efficiency of the method is evaluated by the number of steps or actions taken in one successful run. We then compute the average and standard deviation of those statistics over 1000 tests $\times$ 30 runs of each setup.

\section{Results and Discussion}

\subsection{Q-value updating behaviour}

First, we evaluate the training process by inspecting the behaviour of the DE-QT time series. Figure \ref{fig:entropy} depicts the changes in the sum of entropy values of different flag states over training episodes, averaged over 30 runs. The changes in entropy values of local flag state representations and global state representations with 1, 2 or 3 flags in training reach a steady state in less than 2000 episodes, much faster than the decay of entropy values in compact global state setup (3-states setup).

Particularly, in the case of \emph{Global-1-8}, the entropy values of both flag states experience a sudden drop before 1000 episodes are reached. This behaviour represents a fast information loss compared to other setups. Interestingly, the decrease of entropy values is less rapid when the number of flags in training increases. The least information loss at each flag state is achieved in \emph{Global-8-8} (8 flags in training). The higher the dimension of flag states, the more the tensor is able to maintain information over time. Thus, the loss of information at each state become smaller, which maintains the performance better when training prolongs.

In the case of local state representations, the behaviour of entropy drops faster at a very early stage of training. Further analysis does not support that local state representations help the Q-table converge to a near optimal solution.

\subsection{Generalization test results}

After training with different setups with dynamic flag positioning, the final Q-table after 10000 episodes in each setup is used to test on a \enquote{general} situation in which there are flags located anywhere within the flag zone. The higher the number of flags collected, the better generalization the agent\textquoteright s learned policy achieves. 

The success of one setup depends on the success rate when the agent collects all 8 flags and proceeds to the goal, and the efficiency represented by the number of steps taken in successful runs. The mean discounted reward reflects the general performance of the agent regarding both success rate and efficiency.

Table \ref{table:test-statistics} compares the test results of different flag state representations. In the tests at $t_{10000}$ with global state space representations, when increasing the number of flags in training, the discounted reward, number of flags collected and the success rate improve. At Global-8-8, the success rate is nearly 100\%. The phenomenon can be explained with the increase of the probability that the agent has encountered flags when training with higher number of flags. However, the mean rewards of all setups are lower than 6.00 due to low efficiency.

In relation to the entropy analysis, the results obtained from the testing phase suggests that the higher dimension in flag state of global state representations, such as \emph{Global-7-8} and \emph{Global-8-8} and the compact-global state representations maintain a high DE value. Thus, the agent avoids losing valuable information too fast under the uncertainty of the environment.

\begin{table*}[t]
\centering
\bgroup
\def\arraystretch{1.5} 
\small{
\begin{tabular}{p{5cm} >{\centering\arraybackslash}m{2cm} >{\centering\arraybackslash}m{3cm} >{\centering\arraybackslash}m{2cm} >{\centering\arraybackslash}m{3cm} }
\hline
\textbf{Setup} & \textbf{Discounted Reward} & \textbf{Number of Flags Collected} & \textbf{Success Rate} & \textbf{Number of Steps in Successful Tests} \\
\hline
 & $\mu \pm \sigma$ & $\mu \pm \sigma$ & $\mu \pm \sigma$ & $\mu \pm \sigma$ \\
\hline
\multicolumn{5}{c}{\textbf{At 10000 Episodes ($t_{10000}$)}}
\\
\hline
\textbf{Global-1-8} & 3.25$\pm$2.45 & 7.57$\pm$1.17 & 0.67$\pm$0.01 & 537.43$\pm$245.32 
\\
\textbf{Global-2-8} & 3.60$\pm$2.38 & 7.69$\pm$0.97 & 0.74$\pm$0.01 & 521.97$\pm$245.52
\\
\textbf{Global-3-8} & 3.64$\pm$2.33 & 7.67$\pm$0.95 & 0.75$\pm$0.04 & 527.87$\pm$241.95 
\\
\textbf{Global-4-8} & 3.42$\pm$2.29 & 7.48$\pm$1.29 & 0.73$\pm$0.04 & 560.38$\pm$236.67 
\\
\textbf{Global-5-8} & 3.88$\pm$2.24 & 7.48$\pm$1.46 & 0.79$\pm$0.10 & 514.92$\pm$234.89 
\\
\textbf{Global-6-8} & 4.71$\pm$1.89 & 7.76$\pm$1.11 & 0.90$\pm$0.07 & 449.65$\pm$225.07 
\\
\textbf{Global-7-8} & 5.69$\pm$1.23 & \textbf{7.99$\pm$0.27} & 0.99$\pm$0.01 & 344.25$\pm$191.69 
\\
\textbf{Global-8-8} & 6.00$\pm$0.98 & 8.00$\pm$0.05 & 1.00$\pm$0.00 & 297.60$\pm$165.04 
\\
\textbf{Compact Global (3-Flag States)} & 5.59$\pm$1.21 & 7.98$\pm$0.23 & \textbf{0.99$\pm$0.00} & 360.93$\pm$190.26 
\\
\textbf{Local-1-8} & 2.71$\pm$2.09 & \textbf{6.17$\pm$1.93} & \textbf{0.19$\pm$0.02} & 560.66$\pm$250.24 
\\
\textbf{Local-8-8} & 2.22$\pm$2.51 & \textbf{7.40$\pm$1.27} & 0.26$\pm$0.17 & 497.19$\pm$250.71 
\\ 
\hline
\hline
\multicolumn{5}{c}{\textbf{Early stopping at highest sum of differential entropy values ($t_{max}$)}}
\\
\hline

\textbf{Global-1-8} & \textbf{6.60$\pm$0.77} & \textbf{7.82$\pm$0.77} & \textbf{0.91$\pm$0.05} & \textbf{177.25$\pm$93.21} 
\\
\textbf{Global-2-8} & \textbf{6.71$\pm$0.87} & \textbf{7.92$\pm$0.41} & \textbf{0.95$\pm$0.05} & \textbf{168.28$\pm$107.46} 
\\
\textbf{Global-3-8} & \textbf{6.42$\pm$1.05} & \textbf{7.93$\pm$0.44} & \textbf{0.97$\pm$0.03} & \textbf{217.81$\pm$146.18} 
\\
\textbf{Global-4-8} & \textbf{5.90$\pm$1.77} & \textbf{7.76$\pm$0.94} & \textbf{0.93$\pm$0.15} & \textbf{259.17$\pm$185.18} 
\\
\textbf{Global-5-8} & \textbf{5.75$\pm$2.07} & \textbf{7.57$\pm$1.39} & \textbf{0.91$\pm$0.17} & \textbf{255.05$\pm$197.89} 
\\
\textbf{Global-6-8} & \textbf{6.47$\pm$1.48} & \textbf{7.89$\pm$0.99} & \textbf{0.97$\pm$0.08} & \textbf{193.00$\pm$161.96} 
\\
\textbf{Global-7-8} & \textbf{6.49$\pm$1.59} & 7.76$\pm$1.27 & 0.96$\pm$0.10 & \textbf{182.42$\pm$164.15}
\\
\textbf{Global-8-8} & \textbf{7.18$\pm$0.41} & 8.00$\pm$0.03 & 1.00$\pm$0.00 & \textbf{110.68$\pm$60.96} 
\\
\textbf{Compact Global (3-Flag States)} & \textbf{6.57$\pm$1.18} & 7.92$\pm$0.27 & 0.92$\pm$0.07 & \textbf{172.57$\pm$120.11} 
\\
\textbf{Local-1-8} & \textbf{3.06$\pm$1.84} & 5.53$\pm$1.95 & 0.14$\pm$0.03 & \textbf{540.01$\pm$249.81} 
\\
\textbf{Local-8-8} & \textbf{2.47$\pm$2.51} & 7.31$\pm$1.39 & \textbf{0.28$\pm$0.21} & 493.05$\pm$246.42 
\\ 
\hline
\end{tabular}
}
\egroup
\caption{Statistics of 30 runs $\times$ 1000 tests of two testing cases for different state space representations. The figures in bold are significantly better than their counterparts in the other test (t-test is performed at a significant level of 0.05).}
\label{table:test-statistics}
\end{table*}

\subsection{Effects of early stopping}

The early testing examines our hypothesis whether the performance of an RL agent decays while the amount of valuable information decreases over time due to an over-training problem.

When taking Q-tables in maximum-sum-of-entropy setting in early stopping test (at $t_{max}$), the cumulative discounted reward the agent obtains on average is significantly higher than the performance obtaining at $t_{10000}$ from about 17\% up to about 200\%. Most success rates when testing at $t_{max}$ is significantly better than success rates at $t_{10000}$. There are no significant difference between these figures at $t_{max}$ and $t_{10000}$ in \emph{Global-7-8}, \emph{Global-8-8}, and \emph{Compact Global} setups, but the efficiency when testing at $t_{max}$ is superior.

In relation to the differential entropy values of the Q-tables, less information loss when overtraining the agent with higher number of flags in training leads to the preservation of success and an increase in the number of flags collected on average. However, early stopping has a great impact on the efficiency of the Q-learning algorithm verified by the lower number of steps to reach the goal in successful runs. 

The boxplots in Figure \ref{fig:boxplot} suggests that in most flag state representations, the performance is lower when testing earlier than the time $t_{max}$ due to under-training problem. The success rate at later stage might be lower, equivalent or even higher than tests' performance at $t_{max}$. However, the efficiency reduces over time. There is a trade-off between high success rate and high efficient when we attempts to choose an appropriate stopping point for training sessions. The change in DE-QT time-series is useful in information loss detection. Choosing Q-tables at a peak of DE-QT guarantees a high success rate while maintaining high efficiency.

\section{Conclusion}

State space representations directly affect the performance of a Q-learning algorithms. The choice of a state space representation have been investigated in a variety of research papers using both methods of manual design and automatic selection of state space representation combined with Q-learning. The performance of a RL agent, however, is conventionally assessed through the reward.

In this paper, we propose a novel method computing DE-QT, which can reveal the change in information of the Q-table. Examining the learning process offers an insight into the evolution of the agent's knowledge during skill acquisition.

The task used in our demonstration is a flag collection task - a gridworld problem adapting the dynamic dispositions of objects in the environment. Differential entropy values have been used as an indicator of generalization success of a state space representation. It offers a way to choose an appropriate stopping point in the training process.

The scope of this paper only focuses on investigating the feasibility of the information loss detection method for the success or failure of state space representations. Future directions will extend the investigation to other RL problem.

%

\clearpage{}
\bibliography{references}

\begin{thebibliography}{}

\bibitem[\protect\citeauthoryear{Araabi, Mastoureshgh, and
  Ahmadabadi}{2007}]{araabi2007study}
Araabi, B.~N.; Mastoureshgh, S.; and Ahmadabadi, M.~N.
\newblock 2007.
\newblock A study on expertise of agents and its effects on cooperative $ q
  $-learning.
\newblock {\em IEEE Transactions on Systems, Man, and Cybernetics, Part B
  (Cybernetics)} 37(2):398--409.

\bibitem[\protect\citeauthoryear{Dearden, Friedman, and
  Russell}{1998}]{dearden1998bayesian}
Dearden, R.; Friedman, N.; and Russell, S.
\newblock 1998.
\newblock Bayesian q-learning.
\newblock In {\em AAAI/IAAI},  761--768.

\bibitem[\protect\citeauthoryear{Kaelbling, Littman, and
  Moore}{1996}]{kaelbling1996reinforcement}
Kaelbling, L.~P.; Littman, M.~L.; and Moore, A.~W.
\newblock 1996.
\newblock Reinforcement learning: A survey.
\newblock {\em Journal of artificial intelligence research} 4:237--285.

\bibitem[\protect\citeauthoryear{Kober, Bagnell, and
  Peters}{2013}]{kober2013reinforcement}
Kober, J.; Bagnell, J.~A.; and Peters, J.
\newblock 2013.
\newblock Reinforcement learning in robotics: A survey.
\newblock {\em The International Journal of Robotics Research}
  32(11):1238--1274.

\bibitem[\protect\citeauthoryear{Kormushev, Calinon, and
  Caldwell}{2010}]{kormushev2010robot}
Kormushev, P.; Calinon, S.; and Caldwell, D.~G.
\newblock 2010.
\newblock Robot motor skill coordination with em-based reinforcement learning.
\newblock In {\em Intelligent Robots and Systems (IROS), 2010 IEEE/RSJ
  International Conference on},  3232--3237.
\newblock IEEE.

\bibitem[\protect\citeauthoryear{Lee and Lau}{2004}]{lee2004adaptive}
Lee, I.~S., and Lau, H.~Y.
\newblock 2004.
\newblock Adaptive state space partitioning for reinforcement learning.
\newblock {\em Engineering applications of artificial intelligence}
  17(6):577--588.

\bibitem[\protect\citeauthoryear{Mannor, Rubinstein, and
  Gat}{2003}]{mannor2003cross}
Mannor, S.; Rubinstein, R.~Y.; and Gat, Y.
\newblock 2003.
\newblock The cross entropy method for fast policy search.
\newblock In {\em Proceedings of the 20th International Conference on Machine
  Learning (ICML-03)},  512--519.

\bibitem[\protect\citeauthoryear{Mao, Cheng, and Ray}{2012}]{mao2012q}
Mao, T.; Cheng, Z.; and Ray, L.~E.
\newblock 2012.
\newblock Q-tree: automatic construction of hierarchical state representation
  for reinforcement learning.
\newblock In {\em International Conference on Intelligent Robotics and
  Applications},  562--576.
\newblock Springer.

\bibitem[\protect\citeauthoryear{Mitsunaga \bgroup et al\mbox.\egroup
  }{2008}]{mitsunaga2008adapting}
Mitsunaga, N.; Smith, C.; Kanda, T.; Ishiguro, H.; and Hagita, N.
\newblock 2008.
\newblock Adapting robot behavior for human--robot interaction.
\newblock {\em IEEE Transactions on Robotics} 24(4):911--916.

\bibitem[\protect\citeauthoryear{Murao and Kitamura}{1997}]{murao1997q}
Murao, H., and Kitamura, S.
\newblock 1997.
\newblock Q-learning with adaptive state segmentation (qlass).
\newblock In {\em Computational Intelligence in Robotics and Automation, 1997.
  CIRA'97., Proceedings., 1997 IEEE International Symposium on},  179--184.
\newblock IEEE.

\bibitem[\protect\citeauthoryear{Peters, M{\"u}lling, and
  Altun}{2010}]{peters2010relative}
Peters, J.; M{\"u}lling, K.; and Altun, Y.
\newblock 2010.
\newblock Relative entropy policy search.
\newblock In {\em AAAI},  1607--1612.
\newblock Atlanta.

\bibitem[\protect\citeauthoryear{Sutton and
  Barto}{1998}]{sutton1998reinforcement}
Sutton, R.~S., and Barto, A.~G.
\newblock 1998.
\newblock {\em Reinforcement learning: An introduction}.
\newblock MIT press Cambridge.

\bibitem[\protect\citeauthoryear{Watkins and Dayan}{1992}]{Watkins1992}
Watkins, C. J. C.~H., and Dayan, P.
\newblock 1992.
\newblock Q-learning.
\newblock {\em Machine Learning} 8(3):279--292.

\end{thebibliography}
\bibliographystyle{aaai}
\end{document}